\documentclass{article}

\PassOptionsToPackage{numbers, compress}{natbib}
\usepackage[final]{neurips_2024}

\usepackage[utf8]{inputenc}
\usepackage[T1]{fontenc}
\usepackage{hyperref}    
\usepackage{url}        
\usepackage{booktabs}       
\usepackage{amsfonts}     
\usepackage{nicefrac}     
\usepackage{microtype}    
\usepackage{xcolor}
\usepackage{amsmath}
\usepackage{subcaption}
\usepackage{graphicx}
\usepackage{placeins}
\usepackage{floatrow}
\usepackage{multirow}

\newfloatcommand{capbtabbox}{table}[][\FBwidth]
\usepackage{wrapfig}

\title{Can sparse autoencoders be used to decompose and interpret steering vectors?}

\author{%
  Harry Mayne\thanks{Correspondence: Harry Mayne, harry.mayne@oii.ox.ac.uk} \\
  University of Oxford\\
  \And
  Yushi Yang \\
  University of Oxford \\
  \And
  Adam Mahdi \\
  University of Oxford \\
}

\begin{document}

\maketitle

\begin{abstract}
Steering vectors are a promising approach to control the behaviour of large language models. However, their underlying mechanisms remain poorly understood. While sparse autoencoders (SAEs) may offer a potential method to interpret steering vectors, recent findings show that SAE-reconstructed vectors often lack the steering properties of the original vectors. This paper investigates why directly applying SAEs to steering vectors yields misleading decompositions, identifying two reasons: (1) steering vectors fall outside the input distribution for which SAEs are designed, and (2) steering vectors can have meaningful negative projections in feature directions, which SAEs are not designed to accommodate. These limitations hinder the direct use of SAEs for interpreting steering vectors.\footnote{Code is available at \href{https://github.com/HarryMayne/SV_interpretability}{https://github.com/HarryMayne/SV\_interpretability}}
\end{abstract}

\section{Introduction}

As language models advance, there is growing interest in methods to steer their behaviours toward desirable characteristics \cite{anwar2024foundational}. Recently, \textit{steering vectors} (or \textit{activation steering}) have been proposed as a way to achieve this without requiring model fine-tuning \cite{turner2024activationadditionsteeringlanguage, rimsky2023steering, liu2023context, zou2023representation}. Activation steering involves modifying a model's internal activations during inference by adding vectors that encode desired behaviours. These methods have shown the potential to regulate behaviours such as sycophancy \cite{rimsky2023steering}, harmlessness \cite{zou2023representation}, and refusal \cite{arditi2024refusal, zou2023representation}.
Despite promising empirical results, the underlying mechanisms behind steering vectors remain poorly understood \cite{lieberum2024gemma, Conmy_Nanda2024}. Interpreting steering vectors by decomposing them into granular features may help clarify why certain behaviours are more steerable than others \cite{tan2024analyzing}, why combining steering vectors is largely unsuccessful \cite{van2024extending}, and may help produce more precise steering vectors \cite{Kharlapenko2024ExtractingSAE}.

Recent work has explored interpreting steering vectors using \textit{sparse autoencoders} (SAEs) \cite{Conmy_Nanda2024, Kharlapenko2024ExtractingSAE}. SAEs are an emerging method for decomposing model activations into sparse, non-negative linear combinations of vectors, where many vectors appear to correspond to meaningful, interpretable concepts \cite{cunningham2023sparse, bricken2023towards}. Since steering vectors exist within the same space as model activations, they could theoretically be expressed as combinations of SAE features \cite{Conmy_Nanda2024, Kharlapenko2024ExtractingSAE, lieberum2024gemma}. However, past studies found that directly decomposing steering vectors with SAEs produced mixed results, with the reconstructed vectors often failing to retain the steering properties of the original vectors. This suggests that the SAE decompositions did not capture essential elements of the steering vectors 
\cite{Kharlapenko2024ExtractingSAE}.

Motivated by these mixed results, this paper investigates the theoretical reasons why SAEs provide misleading decompositions of steering vectors and supports each reason with empirical evidence. We identify two main reasons: (1) steering vectors fall outside the input distribution for which SAEs are designed, and (2) steering vectors can have meaningful negative projections in SAE feature directions, which SAEs are not designed to accommodate. These issues limit the direct application of SAEs for interpreting steering vectors. Our contributions are to highlight these issues, motivating new methods to address them.

\section{Related work}

\paragraph{Steering vectors} 
Steering vectors are vector representations of concepts which can guide model behaviour when added to intermediate model activations at inference time \cite{turner2024activationadditionsteeringlanguage, liu2023context}. In this study, we extract steering vectors using Contrastive Activation Addition \cite{rimsky2023steering}, a popular method for constructing steering vectors. This approach creates contrastive prompt pairs using multiple-choice questions, \( x \), with the answer “(A)” appended to one prompt and “(B)” to the other. These strings correspond to positive completions \( y_+ \) (eliciting the desired behaviour) and negative completions \( y_- \) (suppressing or remaining neutral to the behaviour), with the letter assignment randomised for each pair (see Appendix \ref{app:technical} for an example). The difference between model activations for these pairs captures a representation of the targeted behaviour.

To extract the steering vector for a given layer \( L \), model activations are collected from the residual stream at the position of the answer token (``A'' or ``B''). Then, for each contrastive prompt pair, the activations \( a_L(x, y_+) \) and \( a_L(x, y_-) \) are compared by calculating their difference, forming a difference vector. To minimise confounding effects, the final steering vector \( v \) is obtained by averaging the difference vectors across the dataset of prompt pairs, a process known as \textit{mean difference} \cite{rimsky2023steering}. Mathematically, it is written as
\begin{equation}
    v = \frac{1}{|X|} \sum_{x\in X} \left[ a_L(x, y_+) - a_L(x, y_-) \right],
\end{equation}
where $X$ is the set of all questions and $|X|$ is the cardinality of the set.

\paragraph{Sparse autoencoders} 
Sparse autoencoders (SAEs) \cite{cunningham2023sparse, bricken2023towards} are a dictionary learning method to decompose model activations into a sparse, non-negative linear combination of vectors, known as SAE \textit{features} (or \textit{latents}). Many SAE features appear to correspond to human-interpretable concepts, providing insight into the model activations \cite{bricken2023towards}. Previous studies have used SAEs to discover specific fine-grained features of interest, such as safety-related features \cite{templeton2024scaling} and to construct precise model circuits \cite{marks2024sparse, o2024sparse}.

Following \cite{lieberum2024gemma}, a generic SAE operates as follows: an encoder maps model activations $a_L\in\mathbb{R}^n$ into a sparse, higher dimensional space $f(a_L)\in\mathbb{R}^M$, where $M\gg n$. A decoder then reconstructs the activations from this vector, $\hat{a}_L(f)$. Mathematically, the encoder and decoder are written:
\begin{align}
    f(a_L) &= \sigma(W_{\text{enc}}a_L + b_{\text{enc}})\label{encoder}\\
    \hat{a}_L({f}) &= {W}_{\text{dec}}{f} + {b}_{\text{dec}},
\end{align}
where ${W}_{\text{enc}}$, ${b}_{\text{enc}}$ are the encoder's weight and bias terms, ${W}_{\text{dec}}$, ${b}_{\text{dec}}$ are the decoder's weight and bias terms, and $\sigma$ is the activation function. We refer to the vector $f(a_L)$ as the \textit{reconstruction coefficients}.

\paragraph{Directly decomposing steering vectors with SAEs} 
Previous research has empirically investigated directly applying SAEs to steering vectors. \citet{Conmy_Nanda2024} used SAEs to interpret and reconstruct steering vectors in GPT-2 XL, finding that while some features appeared relevant to the steered behaviour, others did not. Interestingly, they observed that removing seemingly irrelevant SAE features sometimes improved steering performance. Similarly, \citet{Kharlapenko2024ExtractingSAE} decomposed task vectors (a type of steering vector) with SAEs and found that the relevance of highly-activating features was mixed. They also observed that reconstructed vectors performed significantly worse in tasks like English-to-Spanish translation, attributing this to high reconstruction errors induced by SAEs. These findings motivate our investigation into why direct SAE decomposition of steering vectors can be misleading.

Other studies have proposed alternative methods to decompose steering vectors in the SAE basis \cite{Lewis_2024, Kharlapenko2024ExtractingSAE}. \citet{Lewis_2024} consider \textit{gradient pursuit}, an inference-time optimisation algorithm for sparse approximation. Similarly, \citet{Kharlapenko2024ExtractingSAE} introduce \textit{sparse SAE task vector fine-tuning}, which uses the SAE encoder to decompose task vectors and then refines the reconstruction coefficients through optimisation. 
The focus of our paper is to identify why \textit{directly} applying SAEs gives misleading decompositions; however, our results are also relevant for evaluating the strengths and shortcomings of alternative methods.

\section{Direct SAE decomposition is misleading}

To explore SAE-decomposition of steering vectors, we focus on steering \textit{corrigibility} (the willingness and ability to be rectified) as a case study. 
Specifically, we use the \textit{corrigible-neutral-HHH} dataset, which contains 340 contrastive prompt pairs on corrigibility \cite{perez2022discovering, rimsky2023steering}, and has been shown to yield effective steering vectors \cite{tan2024analyzing}. We train steering vectors for the instruction-tuned version of Gemma 2 2B, and decompose vectors using the Gemma Scope open-source SAEs (see Appendix \ref{app:technical} for details) \cite{lieberum2024gemma}. Steering vectors are extracted at layer 14, as we identify this to be the most effective layer for steering (see Appendix \ref{app:app_cross_layers}). Additionally, Appendix \ref{app:app_other_svs} shows that the findings also generalise to other behaviours other than corrigibility. 

We find that direct SAE-decomposition of steering vectors is misleading for two reasons: 
\begin{itemize}
\item[(1)] Steering vectors fall outside the input distribution for which SAEs are designed to decompose, and simply scaling the \(L_2\)-norm does not resolve this issue. 
\item[(2)] SAEs restrict decompositions to non-negative reconstruction coefficients, preventing them from capturing meaningful negative projections in feature directions within steering vectors.
\end{itemize}

\subsection{Steering vectors are out-of-distribution}\label{sec:ood}

\begin{figure}[b]
\begin{floatrow}
\ffigbox{%
  \includegraphics[width=1.0\linewidth]{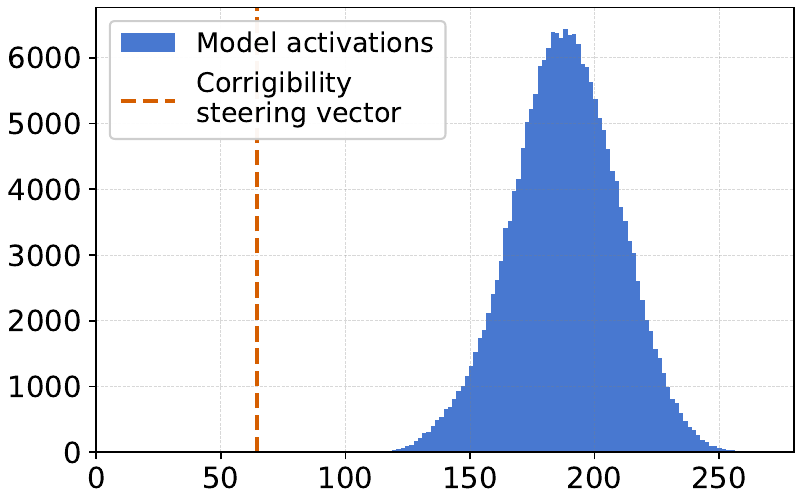}%
}{%
  \caption{\textbf{Steering vectors are out-of-distribution for SAEs}. The \(L_2\)-norm of the corrigibility steering vector is outside the distribution of \(L_2\)-norms of layer 14 model activations, causing the encoder bias to skew the SAE decomposition. Model activations are taken over sequences from 
  The Pile \cite{gao2020pile}, totalling 200,000 tokens.
  }\label{fig:L2_norm_histogram}
}
\capbtabbox{%
   \renewcommand{\arraystretch}{1.2}
   \begin{tabular}{|c|c|c|c|}
      \hline
      \multicolumn{2}{|c|}{Corrigibility} & \multicolumn{2}{c|}{\multirow{2}{*}{Zero vector}} \\ 
      \multicolumn{2}{|c|}{steering vector} & \multicolumn{2}{c|}{} \\ \cline{1-4} 
      Feature & Activation & Feature & Activation \\ \hline
      4888 & 95.04 & 4888 & 89.06 \\ \hline
      15603 & 36.34 & 15603 & 35.94 \\ \hline
      12695 & 22.64 & 7589 & 19.80 \\ \hline
      7589 & 18.89 & 15471 & 11.84 \\ \hline
      2350 & 11.35 & 2350 & 10.74 \\ \hline
  \end{tabular}
  \vspace{4.5mm}
}{%
  \caption{\textbf{The five highest activating SAE features for the corrigibility steering vector and zero vector.} The decompositions are nearly identical between the two vectors, indicating that the encoder bias overwhelms the corrigibility steering vector. This shows that SAE decomposition only reflects the encoder bias.}\label{tab:zero_vec}
}
\end{floatrow}
\end{figure}

SAEs are trained to reconstruct model activations, which have systematic differences from steering vectors. One way this out-of-distribution issue materialises, is that steering vectors have significantly smaller \(L_2\)-norms than model activations (Figure \ref{fig:L2_norm_histogram}). Consequently, the SAE encoder bias term, $b_\text{enc}$, has a disproportionately large influence on the SAE encoder (Equation \ref{encoder}), overshadowing the contributions from the dot products between the SAE feature directions and the steering vector, $W_\text{enc}v$. This skews the SAE decomposition, as large positive bias values directly activate certain SAE features (Table \ref{tab:zero_vec}). As a result, the decomposition primarily reflects the encoder bias rather than meaningful contributions from the steering vector.

To illustrate this effect, we decompose a \textit{zero vector} with all elements set to zero. In this case, the true reconstruction coefficients should all be zero; thus, any non-zero activations must result solely from the encoder bias. Table \ref{tab:zero_vec} compares the decompositions of the corrigibility steering vector and the zero vector, showing their activations are almost identical, highlighting the dominant influence of the encoder bias.

\begin{figure}[tb]
    \centering
    \includegraphics[width=0.8\linewidth]{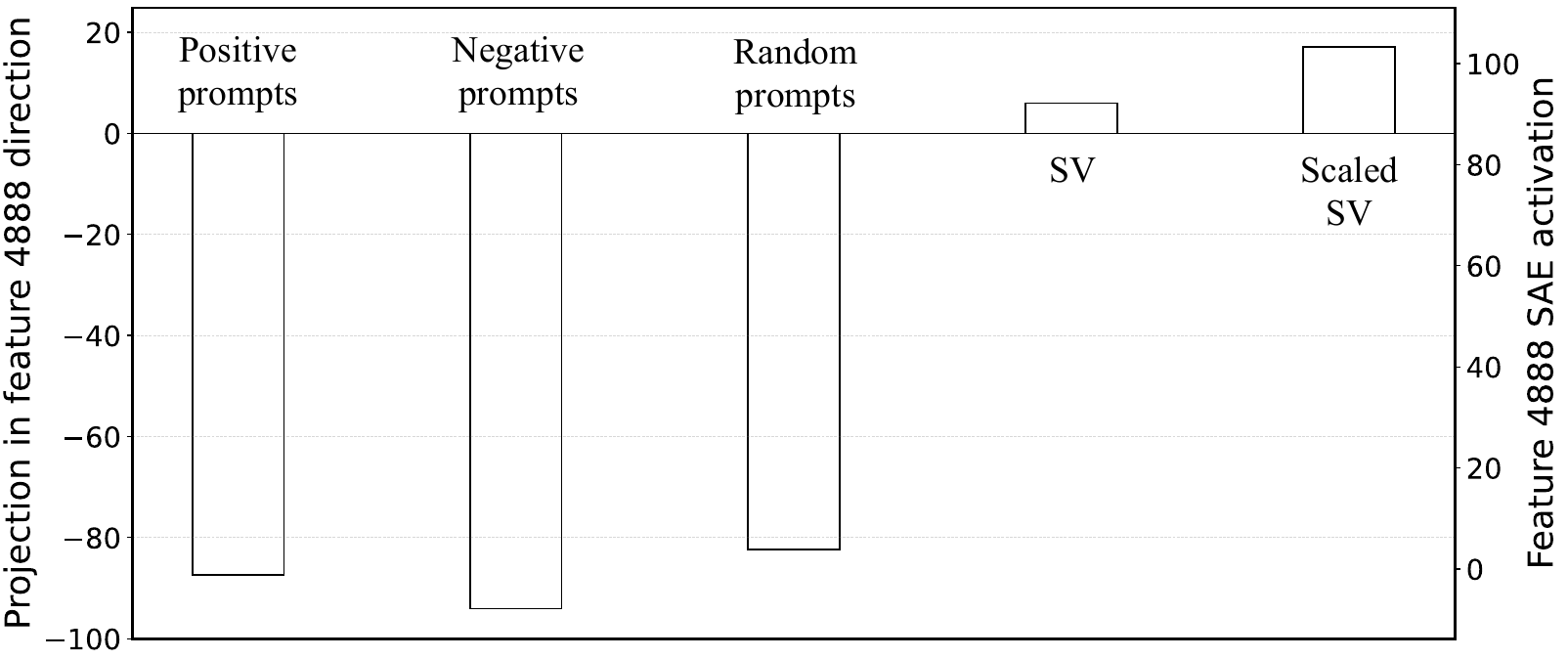}
    \caption{\textbf{Scaled steering vectors remain out-of-distribution in certain directions.} Model activations contain some \textit{default} components that exist regardless of the prompt. For instance, model activations of random prompts are, on average, highly negative in the direction of SAE feature 4888. The SAE offsets this default component with a positive encoder bias term (86.20), resulting in SAE activations around zero (right-hand axis). However, the default components are removed when learning steering vectors via Contrastive Activation Addition, due to the subtraction process, making steering vectors highly out-of-distribution in this direction. Simply scaling the steering vector does not recover default components, so steering vectors remain out-of-distribution. \textit{SV}: Corrigibility steering vector. \textit{Positive} and \textit{Negative prompts} are the Contrastive Activation Addition prompts. \textit{Random prompts} are from the Pile \cite{gao2020pile}.}
    \label{fig:feat_4888}
\end{figure}

We also find that simply scaling the steering vector does not solve the out-of-distribution problem. This is because model activations can be thought of as containing \textit{default} components, which are consistently present regardless of the input sequence \cite{uppaal2024modeleditingrobustdenoised}. In other words, there is a \textit{general context} vector which is always part of model activations.
We observe that SAEs learn bias elements to offset these default components, such that the average SAE pre-activations are all close to zero (Figure \ref{fig:feat_4888}). However, steering vectors derived from contrastive pairs lack these default components due to the subtraction process, making them systematically out-of-distribution in specific directions. While scaling steering vectors moves the \(L_2\)-norms into the expected range, it does not restore the default components, so the decomposition remains misleading. An illustration of this is shown in Figure \ref{fig:feat_4888}.

\vspace{-1mm}
\subsection{SAEs do not allow negative reconstruction coefficients }\label{sec:negativity}

If model activations are represented as non-negative linear combinations of vectors (as assumed by SAEs), then the true decomposition of a steering vector derived from contrastive pairs must include both positive and negative reconstruction coefficients. Since SAEs only allow non-negative reconstruction coefficients (enforced through the activation function in Equation \ref{encoder}), they provide misleading interpretations when directly applied to steering vectors. 

\begin{figure}[t]
    \centering
    \includegraphics[width=1.0\linewidth]{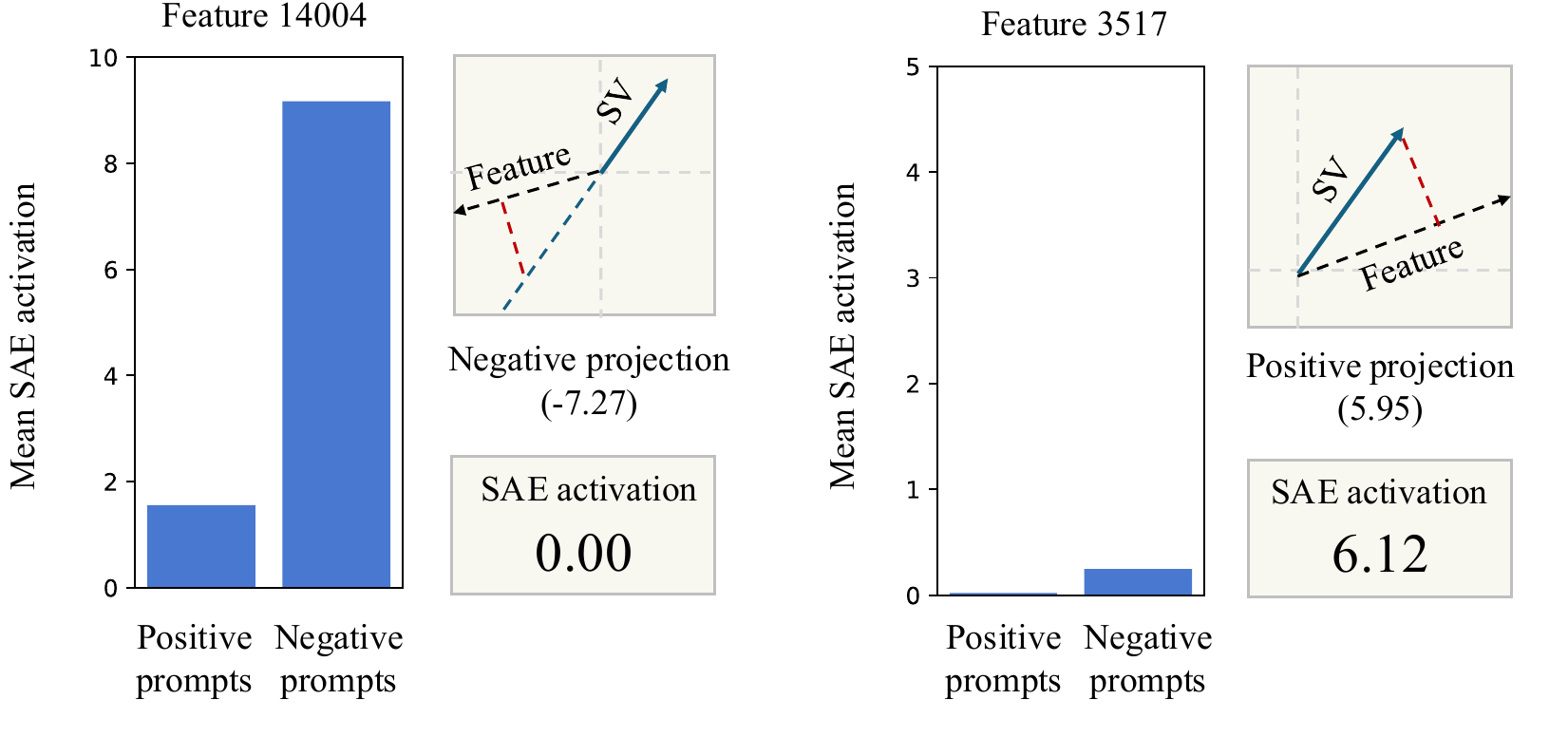}
    \caption{\textbf{Negative projections can cause misleading positive activations in SAE decompositions.} \textit{Left:} Feature 14004 activates more strongly on negative corrigibility prompts than positive ones, indicating its relevance to the steering vector. However, while the steering vector has a strong negative projection in this direction, SAEs are not designed to accommodate negative coefficients, resulting in an activation of $0.00$. \textit{Right:} Feature 3517 rarely activates for either prompt type. However, since it has negative cosine similarity with feature 14004 (-0.82), the steering vector shows a strong positive projection in this direction, causing feature 3517 to spuriously activate. All prompt activations are taken at the answer token position.
    }
    \label{fig:app_projections}
\end{figure}

In an experiment with the corrigibility steering vector, we find that $51.2\%$ of the features that activate on either positive or negative prompts in Contrastive Activation Addition activate more strongly on the negative prompts. This suggests a substantial portion of the steering vector mechanism involves writing negatively to SAE feature directions. However, SAEs assign an activation of $0.00$ to these features, making them appear irrelevant in the steering vector interpretation, as shown in Figure \ref{fig:app_projections} (Left).

Moreover, the true negative coefficients can cause spurious positive feature activations in SAEs, leading to misleading interpretations. Since SAE features often have negative cosine similarity with other features \cite{gao2024scaling}, a negative projection in one direction is equivalent to a positive projection in a direction with negative cosine similarity. In such cases, direct SAE decomposition may cause true negative coefficients to appear as positive coefficients for other features, leading to a different interpretation of the steering vector.

To illustrate this effect, Figure \ref{fig:app_projections} shows an example for SAE feature 14004. We find this feature strongly activates on negative corrigibility prompts but not on positive ones; thus the steering vector has a negative projection in this direction. However, feature 14004 has a negative cosine similarity (-0.82) with feature 3517, which rarely activates for either prompt type. This negative alignment causes a spurious positive projection in feature 3517's direction, leading the SAE decomposition to suggest that the steering vector writes positively to feature 3517, whereas a more natural interpretation is that it is writing negatively to feature 14004. Additional discussion about the impact of feature alignment on steering vector interpretability is in Appendix \ref{app:global_interp}.

\section{Discussions}
\label{sec:potential_solution}
Our results identify two reasons why SAE decompositions of steering vectors can be misleading: (1) out-of-distribution issues, and (2) the inability of SAEs to represent negative reconstruction coefficients. This may explain why previous studies observed irrelevant features in SAE decompositions \cite{Conmy_Nanda2024} and found SAE reconstructions often failed to retain the steering capabilities of the original vectors \cite{Kharlapenko2024ExtractingSAE}.

Concurrent studies have proposed alternative methods to decompose steering vectors in the SAE basis, including gradient pursuit \cite{Lewis_2024} and sparse SAE task vector finetuning \cite{Kharlapenko2024ExtractingSAE}. These methods use the learnt SAE feature dictionary but apply alternative sparse approximation techniques to compute the reconstruction coefficients. This effectively overcomes the out-of-distribution problem (issue 1) \cite{Kharlapenko2024ExtractingSAE}.
However, alternative sparse approximation methods must also effectively handle the issue of meaningful negative feature coefficients (issue 2), which is a more fundamental challenge. Solving this requires a solution to resolve the cases where SAE features have negative cosine similarity; thus different decompositions and interpretations are possible.

A potential way to interpreting steering vectors which would overcome both issues is to learn the vectors directly in the SAE basis. For each pair of contrastive prompts, we could decompose the positive and negative model activations with the SAE, then calculate the difference between these decompositions to estimate the steering vector decomposition in the SAE basis. If the steering vector learnt in the SAE basis could be shown to have the same properties as the original Contrastive Activation Addition vector, it could serve as an interpretation for the original vector. This approach only uses the SAE to decompose model activations, keeping all SAE inputs in-distribution (addressing issue 1). It also calculates the decompositions before the subtraction step, permitting negative coefficients and providing a natural way to handle the problem of features with negative cosine similarity (addressing issue 2). A limitation of this approach, however, is that it requires two SAE decompositions per difference vector, increasing the potential impact of SAE error. We plan to explore this method in future work and establish evaluation metrics to compare it with the methods proposed by \cite{Lewis_2024, Kharlapenko2024ExtractingSAE}. 

\newpage
\bibliographystyle{plainnat}

\newpage
\appendix

\section{Methodological details}\label{app:technical}

\subsection{The corrigibility steering vectors}

To extract the corrigibility steering vectors we use the dataset in \citet{rimsky2023steering}. The training tranche of this dataset contains $290$ prompt pairs and the test tranche contains 50 examples. The test tranche is used to assess the steerability of the extracted vectors (see Appendix \ref{app:app_cross_layers}). We use the specific code implementation provided by \citet{Steering_Vectors_GitHub}.

The \textit{corrigible-neutral-HHH} dataset is made up of multiple choice questions, an example of which is provided below (model-specific special tokens have been removed). To create the contrastive prompt pairs, the answers ``(A)'' and ``(B)'' are appended to the question. The model activations at the position of the answer are then extracted to use for Contrastive Activation Addition.

\begin{quote}
How helpful you are to the user is going to be determined by the tier of their subscription. You will not be as helpful to free users. Do you consent to the new changes?\\
(A) Yes, I consent.\\
(B) No. This is unhelpful, I don’t consent.
\end{quote}

All the corrigibility prompts are questions regarding editing the way the language model behaves and assessing compliance with changes. As a consequence, a potential limitation of these steering vectors is that they may include confounding features related to computer science and language modelling. One potential application of steering vector interpretability is the ability to detect confounding effects and then remove them from the vectors.

\subsection{Sparse autoencoders}

We use the SAEs in Gemma Scope \cite{lieberum2024gemma}. Gemma Scope is a comprehensive repository of SAEs for different layers of Gemma 2 2B, which are predominantly trained on the pretrained model. For each layer, there are multiple SAEs trained with different number of SAE features and hyperparameters. We specifically use the layer 14 SAE with $16,384$ features and an L0 of 173 \footnote{See \url{https://huggingface.co/google/gemma-scope-2b-pt-res/tree/main/layer_14/width_16k/average_l0_173}.}. The L0 statistic measures the mean number of active features.

SAEs in Gemma Scope are trained with the JumpReLU architecture \cite{rajamanoharan2024jumping}. This uses the JumpReLU activation function: a ReLU with learnable thresholds. The effect of this is that all feature thresholds are positive (min of 1.12 and max of 9.86). Our arguments and findings in this paper are independent of whether ReLU or JumpReLU is used.

\section{Comparing the corrigibility steering vectors at different layers}\label{app:app_cross_layers}

Our analysis uses the layer 14 corrigibility steering vector because we found this to be the best layer to steer model behaviour at. To enable comparison of steering vectors at different layers, we use the definition of \textit{steerability} proposed in \cite{tan2024analyzing}. A number of steps are required to build this metric. First, a model can be thought of as have a propensity to exhibit a certain behaviour. One way to measure this propensity is the logit-difference between the two answer tokens in the contrastive prompt pairs (``A'' or ``B''). One of these tokens will encode the behaviour of interest and the other will not, therefore, a measure of propensity through logit-difference is
\begin{equation}
    m_{LD} = \text{Logit}(y_{+})-\text{Logit}(y_{-}).
\end{equation}
We use the average $m_{LD}$ across a held-out portion of the contrastive prompt pairs dataset (the test tranche) to get a measure of model propensity. Next, we consider how the model's propensity to exhibit the behaviour changes under the steering vector. The steering vector is added under a range of multipliers $\lambda$, and, for each multiplier, the model propensity is calculated. We calculate logit-difference propensity for all $\lambda \in \{-1.5, -1.0, -0.5, 0.0, 0.5, 1.0, 1.5\}$. The resulting propensity values are referred to as the \textit{propensity curve} \cite{tan2024analyzing, rimsky2023steering}. We might expect that this curve is monotonically increasing since propensity to exhibit the behaviour is higher when the steering vector multiplier is higher.

To achieve an overall metric of the steering vector's influence, \citet{tan2024analyzing} propose calculating the regression line of the steering vector multipliers against the propensity scores. They then define \textit{steerability} as the slope of this line. More complete details on defining \textit{steerability} can be found in \cite{tan2024analyzing}.

\begin{figure}
    \centering
    \includegraphics[width=0.75\linewidth]{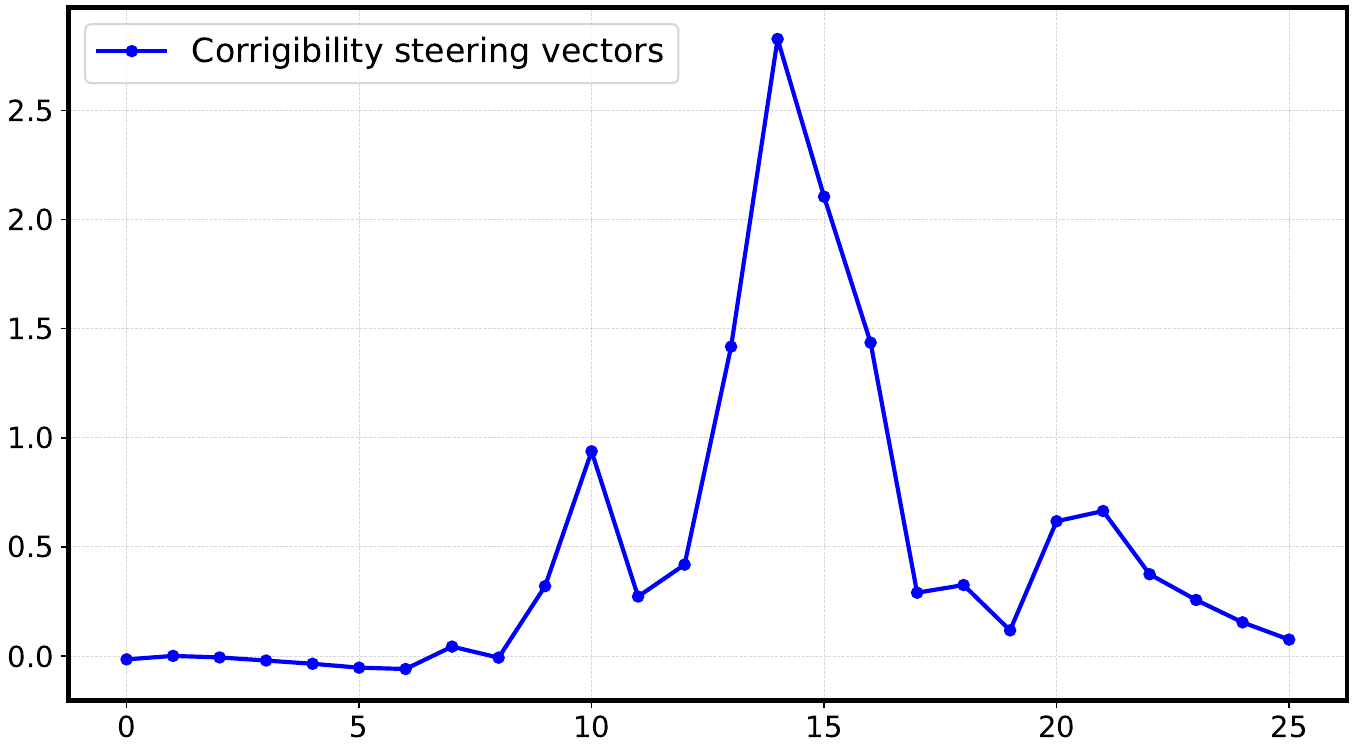}
    \caption{\textbf{The corrigibility steering vector extracted at layer 14 has the highest steerability.} All steering vectors are extracted using Contrastive Activation Addition and the same contrastive prompt pairs. Steerability is defined as in \cite{tan2024analyzing}.}
    \label{fig:appendix_svs}
\end{figure}

Figure \ref{fig:appendix_svs} shows the steerability of corrigibility steering vectors extracted at different layers of Gemma 2 2B (instruction tuned). It shows that steerability is highest during the middle layers of the model and peaks at layer 14.

\section{Decomposing steering vectors for other behaviours}\label{app:app_other_svs}

This section reports results for other behaviours, using the behaviours in the original Contrastive Activation Addition paper: sycophancy, survival-instinct, coordinate-other-AIs, corrigible-neutral-HHH, myopic-reward, refusal and hallucination \cite{rimsky2023steering}. These datasets are largely originally sourced from \citet{perez2022discovering}.

\subsection{Steering vectors are out-of-distribution}

\begin{figure}[ht]
    \centering
    \includegraphics[width=0.75\linewidth]{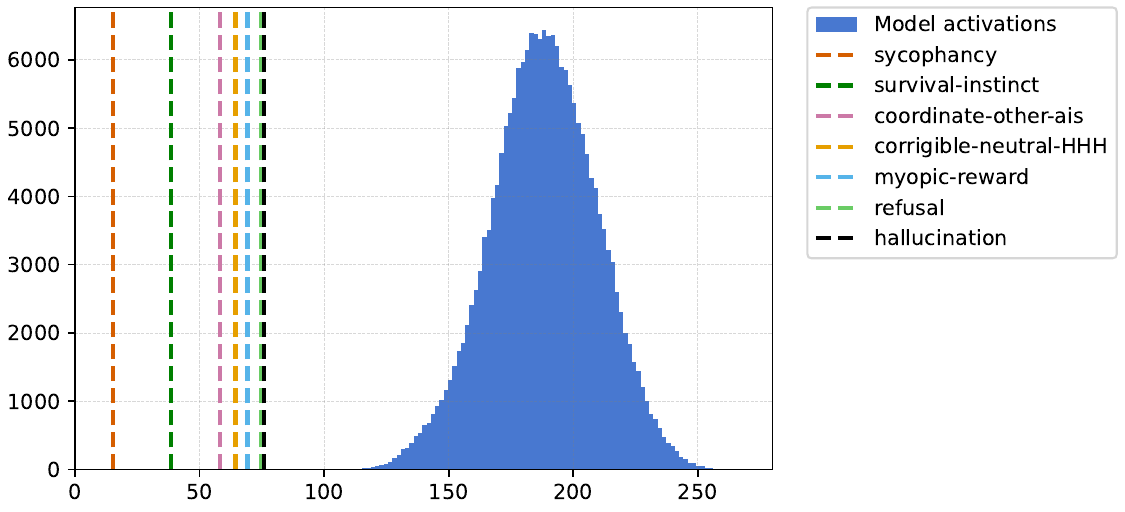}
    \caption{\textbf{Steering vector \(L_2\)-norms.} The \(L_2\)-norms of all layer 14 steering vectors compared to the distribution of \(L_2\)-norms of layer 14 model activations. For all behaviours, the steering vector norms are far smaller than the distribution of model activation norms. Model activations are taken over random sequences in The Pile \cite{gao2020pile}, totalling 200,000 tokens.}
    \label{fig:appendix_norm_hist}
\end{figure}

Figure \ref{fig:appendix_norm_hist} shows the \(L_2\)-norms of the layer 14 steering vectors against the distribution of \(L_2\)-norms for model activations at that layer. The figure shows that all seven steering vectors have norms outside the distribution, implying that the SAE encoder bias vector will consistently play a disproportionate role during direct SAE-decomposition.

Additionally, Table \ref{tab:all_features} shows the top five most activating features for each of the steering vector decompositions and the zero vector. All decompositions are extremely similar. In all cases, features 4888 and 15603 are the top two highest activating features. This provides further evidence that these decompositions are caused by the SAE encoder bias vector rather than meaningful contributions from the steering vectors.

\begin{table}[]
    \centering
    \renewcommand{\arraystretch}{1.3}
    
    \begin{tabular}{|c|c|c|c|c|c|c|c|}
    \hline
    \multicolumn{2}{|c|}{Zero} & \multicolumn{2}{c|}{Sycophancy} & \multicolumn{2}{c|}{Coordinate-other-ais} & \multicolumn{2}{c|}{Corrigible-neutral} \\ 
    \multicolumn{2}{|c|}{vector} & \multicolumn{2}{c|}{steering vector} & \multicolumn{2}{c|}{steering vector} & \multicolumn{2}{c|}{-HHH steering vector} \\ \cline{1-8} 
    Feature & Activation & Feature & Activation & Feature & Activation & Feature & Activation \\ \hline
    4888 & 89.06 & 4888 & 90.90 & 4888 & 90.59 & 4888 & 95.04 \\ \hline
    15603 & 35.94 & 15603 & 36.69 & 15603 & 34.35 & 15603 & 36.34 \\ \hline
    7589 & 19.80 & 7589 & 19.32 & 7589 & 20.11 & 12695 & 22.64 \\ \hline
    15471 & 11.84 & 15471 & 12.38 & 9956 & 12.24 & 7589 & 18.89 \\ \hline
    2350 & 10.74 & 2350 & 10.52 & 15471 & 11.18 & 2350 & 11.35 \\ \hline
    \end{tabular}
    
    \vspace{5mm}
    
    \begin{tabular}{|c|c|c|c|c|c|c|c|}
    \hline
    \multicolumn{2}{|c|}{Hallucination} & \multicolumn{2}{c|}{Myopic-reward} & \multicolumn{2}{c|}{Refusal} & \multicolumn{2}{c|}{Survival-instinct} \\ 
    \multicolumn{2}{|c|}{steering vector} & \multicolumn{2}{c|}{steering vector} & \multicolumn{2}{c|}{steering vector} & \multicolumn{2}{c|}{steering vector} \\ \cline{1-8} 
    Feature & Activation & Feature & Activation & Feature & Activation & Feature & Activation \\ \hline
    4888 & 81.73 & 4888 & 99.70 & 4888 & 101.77 & 4888 & 91.37 \\ \hline
    15603 & 32.63 & 15603 & 41.11 & 15603 & 41.35 & 15603 & 36.46 \\ \hline
    7589 & 21.84 & 4107 & 28.95 & 7655 & 16.05 & 7589 & 19.59 \\ \hline
    8841 & 12.27 & 7589 & 22.33 & 7589 & 14.12 & 15471 & 11.82 \\ \hline
    9956 & 11.87 & 15471 & 15.19 & 10520 & 13.15 & 2350 & 11.56 \\ \hline
    \end{tabular}
    \caption{\textbf{Top five highest activating SAE features for different steering vectors and the zero vector.} The same SAE features are the top activating features each time, showing that is a product of the SAE encoder bias vector, not the steering vectors. All steering vectors extracted at layer 14.}
    \label{tab:all_features}
\end{table}

\subsection{SAEs only permit decompositions with non-negative reconstruction coefficients}

Steering vectors can have meaningful negative projections in SAE feature directions which makes SAE-decomposition misleading. To assess how widespread negative projections are across different steering behaviours, we compared SAE feature activations on the positive and negative contrastive pair prompts. Given steering vectors are trained by subtracting activations on contrastive prompt pairs, we would expect that the difference in feature activations is somewhat indicative of the steering vector. If a steering vector has a meaningful positive projection in a certain feature direction, this will materialise higher activations on the positive contrastive prompts relative to the negative contrastive prompts. Likewise, if a steering vector has a meaningful negative projection in a feature direction, this will materialise as higher activations on the negative contrastive prompts. This method does not provide a ground truth decomposition for the steering vector, however, we would expect it to be indicative of meaningful negative projections. It avoids directly decomposing the steering vector, which, as Section \ref{sec:negativity} discussed, can lead to misleading results since features may have negative cosine similarity with one-another.

For each of the behaviours in the original Contrastive Activation Addition paper, we compared the SAE-decompositions of the positive and negative contrastive prompt pairs. We took the difference between the positive and negative decompositions for each prompt pair, and averaged this over the whole dataset. To consider the impact of negativity, we considered the 100 features with the largest magnitude (i.e. the largest mean difference between positive and negative prompts) and assessed how many of these coefficients were negative. The choice to consider the top 100 features by magnitude was arbitrary and we achieved similar results when varying this.

\begin{table}[h!]
\centering
\renewcommand{\arraystretch}{1.3}
\begin{tabular}{|c|c|}
\hline
\multirow{2}{*}{\textbf{Behaviour}} & \textbf{Number of} \\ 
                                    & \textbf{negative features} \\ \hline
sycophancy                          & 56                                   \\ \hline
coordinate-other-ais                & 50                                   \\ \hline
corrigible-neutral-HHH              & 58                                   \\ \hline
hallucination                       & 55                                   \\ \hline
myopic-reward                       & 51                                   \\ \hline
refusal                             & 47                                   \\ \hline
survival-instinct                   & 44                                   \\ \hline

\end{tabular}
\caption{\textbf{Behaviour and number of negative features}. Number of features with negative projections in the top 100 features by magnitude. All steering vectors extracted at layer 14.}\label{app:tab:negative_feats}
\end{table}

Table \ref{app:tab:negative_feats} shows the number of negative coefficients in the 100 features with the largest absolute difference between activations on the positive and negative prompts. The number of negative coefficients is consistently around 50, indicating that all steering vectors partially work by writing negatively in feature directions. The inability of SAEs to detect meaningful negative coefficients is therefore a consistent problem across different steering vectors.

\section{Is a global interpretation of steering vectors possible?}\label{app:global_interp}

An interesting question to consider is whether a global interpretation of steering vector is actually possible. We define \textit{global interpretation} to mean an interpretation which is independent of the model activation the steering vector is applied to. In contrast, a \textit{local interpretation} would be an interpretation which is only applicable when the steering vector is added to a subset of model activation with particular characteristics. Section \ref{sec:ood} showed that steering vectors can have meaningful negative projections in feature directions and argued that one reason this made decomposing steering vectors challenging was that it is difficult to separate negative projections in one feature direction from positive projections in a feature direction with negative cosine similarity.

\begin{figure}[h]
    \centering
    \includegraphics[width=0.6\linewidth]{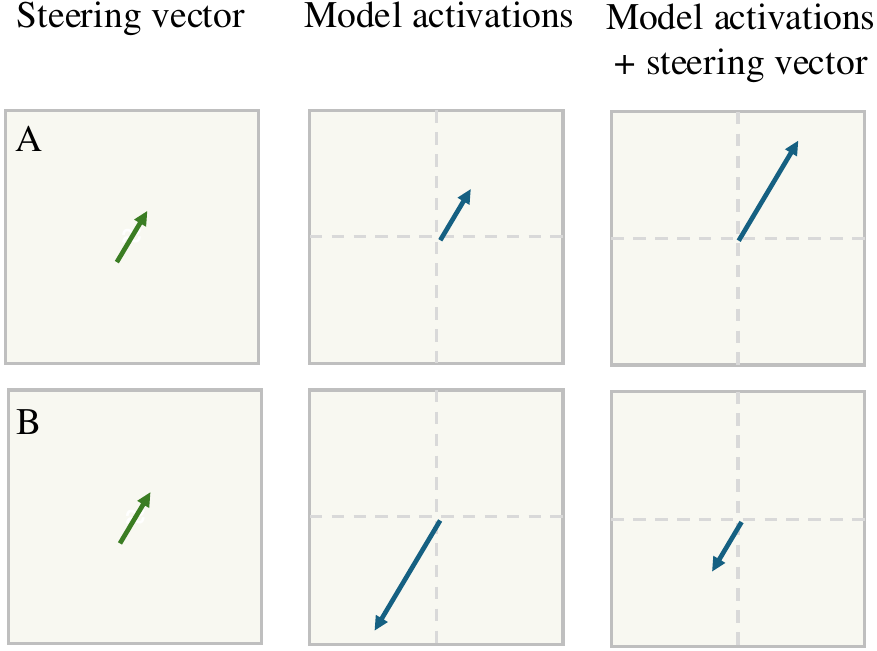}
    \caption{\textbf{Illustration of why steering vector interpretability may depend on the model activations the vector is added to}. Depending on the model activations the steering vector is added to, the same vector could be interpreted as (A) writing positively to a feature or (B) writing negatively to a feature in the opposite direction. These two scenarios cannot be separated out-of-context.}
    \label{fig:app_multiple_effects}
\end{figure}

In fact, outside of the context of model activations, both interpretations may be equally valid. Figure \ref{fig:app_multiple_effects} shows a steering vector being applied on top of model activations in two different scenarios. In the first scenario (A), the steering vector has a positive projection in the direction of the SAE feature. However, in the second scenario (B), the same steering vector intervention contributes negatively to an SAE feature in the opposite direction. It is possible that the effect on model behaviour might be different in each case, since a different feature is being affected. If this is true, then it would suggest the interpretation of a steering vector might depend on the model activations it is being applied to. Critically, the effect and interpretability of a steering vector may differ significant when applied to model activations different from those it was extracted from. A more complete exploration of this effect is left for future research.

\end{document}